\documentclass[lettersize,journal]{IEEEtran}
\usepackage{amsmath,amsfonts}
\usepackage{algorithmic}
\usepackage{algorithm}
\usepackage{array}
\usepackage[caption=false,font=normalsize,labelfont=sf,textfont=sf]{subfig}
\usepackage{textcomp}
\usepackage{stfloats}
\usepackage{url}
\usepackage{verbatim}
\usepackage{graphicx}
\usepackage{booktabs}
\usepackage{multirow}
\DeclareUnicodeCharacter{2212}{\textminus}
\usepackage{subcaption}
\usepackage[utf8]{inputenc}
\usepackage{textgreek}
\usepackage{cite}
\usepackage{tabularray}
\usepackage{rotating}
\usepackage{adjustbox}
\usepackage{soul}
\usepackage[rightcaption,raggedright]{sidecap}
\usepackage{longtable,booktabs}
\usepackage{xcolor}
\begin{document}

\title{Spatiotemporal Context-dependent Personalized Movement Compensation in Delayed Telemanipulation\\}

\author{Sai~Jiang and
        Zonghe~Chua,~\IEEEmembership{Member,~IEEE}
\thanks{Manuscript received Month XX, 2025; revised Month XX, 2025.}
\thanks{This work was supported by ONR Award N00014-23-1-2842.}
\thanks{S. Jiang and Z. Chua are with the Department of ECSE, Case Western Reserve University, Cleveland, OH 44106 USA (e-mail: \texttt{sai.jiang@case.edu}; \texttt{zonghe.chua@case.edu}).}
\thanks{Corresponding author: Sai Jiang.}
}

% The paper headers
% \markboth{Journal of \LaTeX\ Class Files,~Vol.~14, No.~8, August~2021}%
% {Shell \MakeLowercase{\textit{et al.}}: A Sample Article Using IEEEtran.cls for IEEE Journals}

% \IEEEpubid{0000--0000/00\$00.00~\copyright~2021 IEEE}
% Remember, if you use this you must call \IEEEpubidadjcol in the second
% column for its text to clear the IEEEpubid mark.

\maketitle

\begin{abstract}
Communication delay remains a central challenge in telerobotics, where it disrupts visuomotor coordination and reduces task precision. Motion scaling is an effective countermeasure to delay-induced overshoot, yet typical deployments rely on uniform gains that neglect individual and contextual variability. We propose a human-centered method that fits personalized delay-, direction-, and distance-specific scaling parameters for each participant. 

We conducted experiments with twenty participants who performed delayed reaching tasks in a virtual simulator. Scaling gains were computed to minimize mean overshoot in simulation in each combination of experimental conditions. 

%left off here

Evaluation was done in simulation and on a telesurgical robot to evaluate assistance benefits. Performance was assessed across multiple delays, distances, and movement directions using overshoot, endpoint error, trajectory smoothness, economy of motion, and a composite error–time metric. Motion scaling consistently improved performance relative to unassisted trials, yielding up to 20–25\% performance gains in key metrics. Effects were most pronounced at longer delays. Personalization demonstrated additional accuracy benefits for inward reaching at a short distance under moderate delay. The results highlight the potential of personalized scaling as a foundation for more adaptive frameworks that integrate contextual information to improve the safety and precision of teleoperated procedures.
\end{abstract}


\section{Introduction}
\IEEEPARstart{C}{ommunication} delay remains a major challenge in teleoperated surgery \cite{farajiparvarBriefSurveyTelerobotic2020}. . Surgeons have successfully performed transatlantic telesurgery under delays of 155\,ms over a distance of 14,000\,km \cite{marescauxTransatlanticRobotassistedTelesurgery2001a}. However, delays beyond 330\,ms are reported to compromised perceived safety \cite{marescauxTransatlanticRobotassistedTelesurgery2001a}, with latencies exceeding 400\,ms significantly impairing surgeon performance \cite{nahriReviewHapticBilateral2022, nankakuMaximumAcceptableCommunication2022}.

To date, several studies have revealed key principles governing dexterous telemanipulation under delay. These include spatial representation of hypermetria in the motor control system \cite{avrahamStateBasedDelayRepresentation2017}, sub-unity motion scaling compensation strategies \cite{niskyPerceptionActionTeleoperated2011}, and subsequent optimization of scaling gains \cite{limOptimalMotionScaling2025}. While the general benefits of motion scaling and personalization have been demonstrated, the contextual task factors influencing variations in the personalized scaling gain, are still understudied. Within the same individual, delay sensitivity can shift depending on task context, including movement direction, workspace configuration, or whether the task is performed in a virtual simulation or on physical hardware. 
% These observations suggest that delay compensation is not solely a system-level control problem but is also shaped by individual variability and contextual factors.  

In this work, we hypothesized that the performance degradation resulting from delays can be compensated by deriving individualized motion scaling across delay, direction, and reach distance, thereby improving telemanipulation performance relative to non-delayed baseline conditions. To confirm this, we developed a personalized motion compensation approach with fitted parameters across these hypothesized factors. This approach was validated in simulation and tested for its transfer to a real-world manipulation task on the da Vinci Research Kit. Overall, we make the following contributions showing that:
\begin{enumerate}
    \item personalized motion scaling conditioned on delay, distance, and direction improves performance; 
    \item personalized scaling improves accuracy without a reliable increase in movement time across most conditions; 
    \item personalized gains found in simulation transfer to real tasks, but with reduced effectiveness.
\end{enumerate}

\section{Background}
Delayed visual feedback often manifests as overshoot of the intended target, reflecting a mismatch between commanded and perceived motion. Consistent with this signature, Avraham et al. demonstrated that humans perceive delayed feedback not temporally, but rather as a state-based visuomotor gain, manifesting as hypermetria of intended targets \cite{avrahamStateBasedDelayRepresentation2017}. This indicates that delay might be more effectively addressed through spatial scaling adjustments rather than purely temporal corrections.

Their findings supported past results from Nisky et al., who investigated how delays influence perception and action in teleoperated needle insertion into soft tissue \cite{niskyPerceptionActionTeleoperated2011, niskyPerceptuoMotorTransparencyBilateral}. They found that sub-unity motion scaling, where less remote-side movement occurs relative to user input, successfully mitigates delay-induced overshoot without causing perceptual distortions. However, their work was confined to single-degree-of-freedom (DOF) tasks, and it is unclear if these findings generalize to higher-DOF and more complex scenarios.

Sub-unity motion scaling in multi-DOF telemanipulation was most recently investigated  by Richter et al., who explored various motion scaling strategies aimed at improving surgical teleoperation under high-delay conditions \cite{richterMotionScalingSolutions2019a, limOptimalMotionScaling2025}, showing that well-chosen gains can improve accuracy and stability. However, they did not systematically vary delay times or examine how delay interacts with task geometry (direction and distance) to influence performance. 

A number of prior teleoperation delay-mitigation techniques rely on fixed compensation strategies or generic parameters and implicitly assume that all operators are affected by latency in a similar manner\cite{farajiparvarBriefSurveyTelerobotic2020, luAdaptiveControlTime2022, nogueracundarQuantifyingEffectsNetwork2023}. Yet human operators vary in sensorimotor adaptation, which suggests these fixed methods may not fully address individual differences\cite{6657775, dybvikLowcostPredictiveDisplay2021}. 
% Some operators exaggerate their movements under latency, producing larger overshoot, whereas others adopt more conservative control strategies. 
Evidence supporting user-specificity has been shown in the single-DOF scenario, in which personalized gains result in lower hypermetria compared to group-averaged gains \cite{niskyPerceptionActionTeleoperated2011}, and in the multi-DOF scenario \cite{limOptimalMotionScaling2025}, in which a post-hoc optimal gain computed using a probabilistic approach showed variation across different operators. However, the task, perceptual, and behavioral factors that shape the extent of variations are not yet well understood.

\section{Methods}
\noindent The study followed a two-session experimental design. In the first session, referred to as the simulation-only condition (SimOnly), participants performed a reaching task by teleoperating a virtual Patient Side Manipulator (PSM) of a da Vinci Research Kit \cite{kazanzidesOpenSourceResearchKit2014} (dVRK) telesurgical robot using a 6-DOF haptic device and a VR headset. 

After all participants completed the first session, they then returned on a different day for the second session, which tested simulation-to-real (Sim2Real) transfer of the generalized and personalized compensatory motion-scaling gains. They used the same teleoperation interface from the SimOnly session, but instead controlled the PSM of a physical dVRK to perform a peg transfer task. Key phases from the SimOnly session were repeated to evaluate Sim2Real transfer effects. Details of the task design, trial structure, and evaluation metrics are described in Section~\ref{sec:experiment-design}.

\vspace{-1em}
\subsection{Participants}
The study was approved by the Institutional Review Board at Case Western Reserve University. Twenty healthy participants (5 female,  15 male; age range: 18–45 years), all right-handed and without a known history of virtual reality (VR) sickness, motion sickness, or nausea from short-term VR equipment use (less than 30 minutes), were consented and took part in the study. 
% Each participant received a detailed study overview and provided written informed consent prior to participation. 
None of the participants had prior experience with the da Vinci Surgical System or the dVRK, except for 3 participants who were classified as experts as they had experience operating the dVRK through the default Master Tool Manipulators (MTM).

\vspace{-1em}
\subsection{Telemanipulation Hardware Setup}
\begin{figure}
    \centering
    \includegraphics[width=1.0\linewidth]{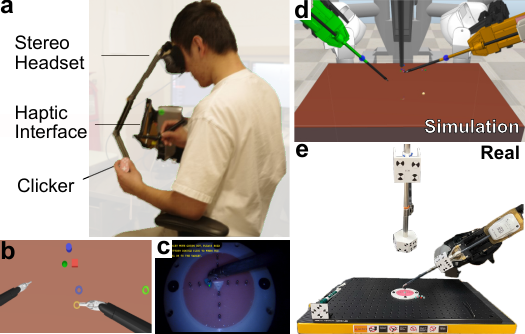}
    \caption{(a) User interface set up. User viewpoints for (b) simulation, (c) real tasks. Follower robot setups in (d) simulation and (e) real scene.}
    
    \label{fig:study setup}
\end{figure}

The experimental setup involved two distinct sessions, as illustrated in Fig.~\ref{fig:study setup}. In the SimOnly session, participants completed a teleoperated shape-matching task within the CoppeliaSim robotic simulator. In the subsequent Sim2Real session, participants performed a peg transfer task using an actual dVRK PSM. 

The SimOnly experimental system was constructed using a kinematic simulation of the PSM within the CoppeliaSim simulator. This virtual environment enabled precise control and consistent replication of robot movements across different experimental conditions. To deliver an immersive 3D experience, participants viewed the virtual scene through a stereoscopic Goovis G3 Max headset, with the virtual Endoscopic Camera Manipulator (ECM) of dVRK providing a stereoscopic viewpoint. The virtual camera parameters in the simulation were calibrated to match those of the physical robot setup, ensuring visual consistency between the SimOnly and Sim2Real sessions.

Participants interacted with the system through a Haply Inverse3 haptic device, which provided 6-DOF motion tracking and 3-DOF force feedback. However, to focus exclusively on translational reaching behavior and avoid confounding effects from wrist orientation, the rotational degrees of freedom were intentionally locked in the virtual PSM. This allowed the study to isolate the effects of delay and compensation strategies on just position control, excluding orientation control.

Both the simulation-only and sim-to-real conditions employed this interface to ensure consistency in user experience and input modalities across sessions. The difference between conditions was the robot being controlled, either the digital twin in simulation or the physical dVRK hardware. The tasks performed in each session were also slightly different: the SimOnly session involved a shape-matching task, while the Sim2Real session involved a modified peg transfer task. 
%This arrangement ensured that any observed differences in performance could be attributed to the change in robot embodiment and task type, rather than variations in input or visual feedback, enabling a controlled evaluation of Sim2Real transfer.

All communication between system components, including the simulation, haptic device, and robot controller, was facilitated through Robot Operating System 2.
% , was handled via the Robot Operating System 2 (ROS2), using the Humble Hawksbill distribution\cite{doi:10.1126/scirobotics.abm6074}. ROS2 facilitated real-time data exchange between the user input, the virtual/real robot arms, and the visualization modules, ensuring consistency and low-latency integration across the experimental setup. 
For the physical robot setup, an optical tracker (MicronTracker, Claronav, Canada) measured the PSM end-effector position  relative to the pegs in the task environment (Fig.\ref{fig:study setup}).

\vspace{-1em}
\subsection{Experimental Design}
\label{sec:experiment-design}
% A single-blind experimental design was used, in which participants were unaware of the presence or absence of communication delay or whether delay compensation was enabled.
The study incorporated four within-subjects factors. In SimOnly, the assistance factor was either Personalized Assistance, or Unassisted. Sim2Real introduced a Generalized Assistance mode as well. The personalized gain was calibrated from each participant’s own baseline, whereas generalized gain was derived from SimOnly aggregates and applied identically to all participants. Assistance was crossed with delay, distance, and direction within participants:
(1) Reaching distance, was the user-side displacement to reach the target. In the SimOnly session, targets were positioned 5.0, 10.0, and 15.0\,mm from the robot’s starting position. With a base motion scaling factor of 0.4, this required user movements of 12.5, 25.0, and 37.5\,mm, respectively. In the Sim2Real session, targets were placed at 10.0, 20.0, and 30.0\,mm, and a 2:1 input scaling was applied, resulting in equivalent user-side displacements. This design ensured consistent motor demands and perceptual reaching distances across both sessions.
(2) Reaching directions were lateral, longitudinal, and vertical.
(3) Communication delays were 100\,ms, 250\,ms, and 400\,ms; based on prior literature identifying 330–500\,ms as the upper bound of acceptable delay in telesurgical tasks.

\vspace{-1em}
\subsection{Task Design}
\begin{figure}
    \centering
    \includegraphics[width=\linewidth]{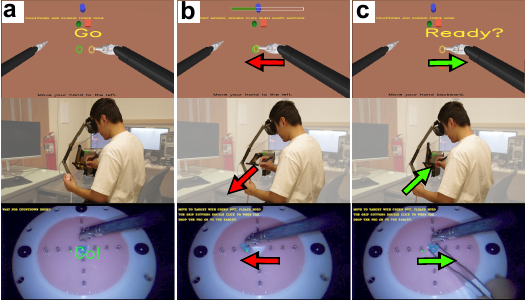}
    \caption{Workflow for a lateral-direction trial. (a) Start of the trial, (b) Movement toward target, (c) reset after completion. Top, middle, and bottom rows show the simulation view, the user interface, and real view, respectively.}
    \label{fig:task-design}
\end{figure}

The task design was adapted from the peg transfer task commonly used as a benchmark in surgical robotics research \cite{SAGES_FLSTasks}. In the SimOnly session, participants performed a shape-matching task, using the haptic device to align a virtual ring with a target ring (Fig.\,\ref{fig:task-design} first row). The target ring was visually presented in the virtual environment, and participants received real-time visual feedback on their alignment performance. 
% Successful task completion required precise translational control of the haptic device to accurately position the ring. 
During the Sim2Real session, participants used the physical dVRK system to complete a modified peg transfer task (Fig.\,\ref{fig:task-design} third row). A ring was initially centered on the pegboard, and participants needed to pick it up and place it onto a target peg.

% \begin{SCfigure*}[][t]
%   \includegraphics[width=.62\textwidth]{figures/task design_v3.pdf}
%   \caption{Workflow for a lateral-direction trial. (a) When ready, the participant presses the button on the haptic device to arm the trial, after which a countdown begins. (b) On the cue, the participant moves toward the target; if the target cannot be reached in a single move, the clutch button is pressed to re-center the hand, and the motion continues. (c) Upon target acquisition, pressing the same button ends the trial; the robot and its simulator reset to the start pose, and the device passively guides the hand back to the start position for the next trial. Red arrows denote user-initiated motion; green arrows denote device-driven motion.}
%   \label{fig:task-design}
% \end{SCfigure*}

For both tasks, participants were instructed to move the haptic device naturally and confidently as they would with their own hand. They were encouraged to complete each trial at a steady and deliberate pace, neither excessively fast nor unnecessarily slow, to promote consistent movements. This approach was intended to elicit participants' predictive control during reaching while minimizing large variations in speed–accuracy trade-offs between participants. Participants were informed that their performance would be evaluated based on both accuracy and completion time.

\vspace{-1em}
\subsection{Procedure}
\subsubsection{Simulation-Only Session}
The simulation-only session consisted of the following phases:

\ul{Phase 0: Consent and System Familiarization.}
Participants were consented, after which the experimenter provided an overview of the study procedure and explained the operation of the haptic input device. An instructional video was shown to demonstrate the task interface and control mechanics.

\ul{Phase 1: Calibration and Practice Block.}
Participants began by adjusting the VR headset’s diopter and interpupillary distance to ensure clear stereoscopic vision. They then adjusted the headset’s height and tilt based on the starting position of the haptic device to maintain a centered posture, with the dominant hand aligned just below chest level. This setup helped standardize forearm configurations, including abduction and adduction across users. They then completed a practice block of 5 trials to familiarize themselves with the system. These trials had random combinations of delay, target distance, and direction. 
%They introduced the key task operations, including how to move and clutch the robot, start, and conclude a trial. 
% For each trial, a combination of delay, target distance, and direction was randomly selected, and a target location was set accordingly. 
At the start of each trial, the haptic device engaged its force feedback to guide the user’s hand to the initial position (Fig.\,\ref{fig:task-design}c). To initiate the trial, the participant double-pressed a button on the device, which triggered a brief on-screen prompt signaling readiness (Fig.\,\ref{fig:task-design}a). As soon as the haptic guidance force disengaged, the participant began reaching toward the target (Fig.\,\ref{fig:task-design}b). When they believed the target had been reached, they double-pressed the same button to end the trial.

\ul{Phase 2: Gain Identification Block.} To minimize the influence of learning effects, each sub-block consisted of five trials: Three delayed trials under experimental conditions (real trials) and two interleaved no-delay catch trials with randomized parameters. These catch trials were embedded between the real trials in each block to reduce predictability. Additionally, one or two extra no-delay catch trials were randomly inserted between sub-blocks. To mitigate fatigue, participants were given a five-minute break after every 20 sub-blocks. The block lasted until each participant performed the task five times per unique trial condition defined by a specific combination of delay, direction, and distance. At its conclusion, participants reported their subjective cognitive and physical workload through the NASA Task Load Index (TLX) survey \cite{hart_development_1988}.

\ul{Phase 3: Baseline Block.} To establish a reference for scaling gain computations, participants completed a block of 56 no-delay trials using the same randomization scheme as in the gain identification (ID) block. Pilot testing showed that collecting the baseline after the gain ID block yielded a more consistent relative baseline overshoot or undershoot for each experimental condition across participants. 
% The data collected in this phase also served as a baseline for evaluating the effectiveness of delay mitigation strategies in later phases.

\ul{Phase 4: Evaluation Block.}
The evaluation phase followed the same structure and condition randomization strategy as the gain ID block, with the difference being that participants experienced personalized compensation gains for each non-delayed trial condition. Each participant performed the task three times per unique condition, defined by a specific combination of delay, direction, and distance. 
% To isolate the effect of compensation strategies, no performance feedback was provided during this phase, and participants were not informed of their accuracy or timing. This ensured that performance metrics reflected the underlying impact of delay compensation rather than continued learning or adaptation. 
Upon completing the evaluation phase, participants once again completed the TLX to assess their perceived cognitive and physical workload. An additional break session was provided after each sub-block to reduce fatigue and maintain performance consistency.

\subsubsection{Sim-to-Real Session}
After all 20 participants had completed the SimOnly session, they were invited back individually to complete the second session (Sim2Real). The overall task structure remained similar to the SimOnly session; however, no gain ID blocks were included. Instead, participants first completed a control block in which they performed the task without delay compensation. The subsequent evaluation phase in the Sim2Real session was divided into two blocks: one using a generic gain (calculated across all users) and the other using the participant’s personalized gain. The order of these two blocks was counterbalanced across participants to mitigate ordering effects.

To reduce the overall trial count and mitigate participant fatigue, two conditions were excluded from the Sim2Real session: the 100\,ms delay and the vertical target direction. The 100\,ms delay was omitted based on prior findings suggesting that short communication delays have limited impact on user perception and task performance \cite{ivanovaShortTimeDelay2021}. The vertical direction was excluded due to physical constraints that made it difficult to replicate the same spatial configuration in the real environment as in the simulation. Furthermore, the real task required participants not only to reach the target but also to physically grasp and place objects onto a peg, introducing greater motor demands and task complexity compared to the simulation. For these reasons, the number of test conditions was intentionally reduced to ensure the session duration remained within a manageable range.

\ul{Phase 1: System Re-familiarization.} Before beginning the Sim2Real session, participants briefly reacquainted themselves with the interface used in the previous session and watched a short instructional video introducing the real-world task. They were then guided on how to teleoperate the physical PSM within the task environment.

\ul{Phase 2: Practice Block.}
After adjusting their posture with the input interface, participants initiated each trial by double-pressing the same button used in the previous session. A visual prompt then appeared on the screen to indicate the start of the trial. A small green dot was displayed to indicate the location of the target peg. Once the guidance force from the haptic device disengaged, participants moved the PSM end-effector vertically downward and pressed the grasp button on the haply to pick up the object. They then transported the object and placed it onto the designated peg. To complete the trial, participants double-pressed the same button again. Upon completion, the haptic device automatically guided their hand back to the starting position, and the PSM returned to its initial configuration.

Due to the increased physical complexity of the real-world task, participants were required to complete twice as many practice trials as in the SimOnly session. This extended familiarization ensured that participants could operate the physical PSM confidently before entering the evaluation phase.

\ul{Phase 3: Control Block.}
The control block in the Sim2Real session was modeled after the gain ID blocks from the SimOnly session, using identical trial composition and condition randomization. Each unique condition was collected three times to serve as a reference for evaluating compensation performance in the subsequent blocks. After completing the control block, participants filled out a brief TLX workload questionnaire.

\ul{Phases 4 \& 5: Evaluation Blocks with Compensation.}
Two evaluation blocks were conducted using different motion scaling strategies: a personalized gain based on each participant’s overshoot behavior from the SimOnly session (Phase 2); a generic gain from aggregated data collected across users. For both blocks, each experimental condition from Phase 2 of the Sim2Real session was repeated five times. After completing the evaluation block, participants again filled out the TLX form.

\vspace{-1em}
\subsection{Derivation of Personalized Compensation Factor}
The gain metric \( G(\delta, d, \theta) \) was a scalar-valued function over a structured three-dimensional input space, where \( \delta \) represents temporal delay, \( d \) denotes the spatial displacement magnitude, and \( \theta \) corresponds to the perturbation direction. Formally,
\begin{equation}
G(\delta, d, \theta) = \frac{\mathbb{E}_{i}[P_0^{(i)}(d, \theta)]}{\mathbb{E}_{j}[P_\delta^{(j)}(d, \theta)]}\text{.}
\end{equation}
Here, \( P_0^{(i)}(d, \theta) \) and \( P_\delta^{(j)}(d, \theta) \) represent the peak reach distances observed in baseline (no-delay) and gain ID (delayed) trials, respectively, under matched spatial conditions. The expectations \( \mathbb{E}_{i} \) and \( \mathbb{E}_{j} \) represent arithmetic mean taken over trials \( i \) and \( j \) within the respective conditions.

This formulation results in a three-dimensional array of scalar gain values indexed by \((\delta, d, \theta)\). Although the structure is not a tensor in the formal algebraic sense, we adopt the term “gain tensor” to reflect its multi-dimensional organization in implementation. Each scalar gain acts as a multiplicative modifier of the original task scaling factor, thereby adjusting the commanded motion according to delay, distance, and direction. This framework facilitates interpolation and cross-condition analysis of the motor system’s delay sensitivity.

%\vspace{-1em}
\subsection{Performance Metrics}
\subsubsection{Initial Reaching Error}

Overshoot, denoted by \( O \), quantifies the maximum displacement of the end-effector beyond the target position along the primary movement axis. For a given trial, let \( x_t \) denote the end-effector position at time \( t \), and \( x_{\text{target}} \) the target position. Then:
\begin{equation}
O = \max_t \left[ (x_t - x_{\text{target}}) \cdot \hat{u} \right],
\end{equation}
where \( \hat{u} \) is the unit vector pointing from the start position to the target. This formulation captures the maximal excursion beyond the target before the motion stabilizes.

In contrast, undershoot, denoted by \( U \), measures how much the trajectory falls short at the first stabilization point near the target. Let \( t^{*} \) be the time index of the first local extremum of the end-effector trajectory along the movement axis, that is, the first time the end-effector stops progressing toward the target. The undershoot is defined as:

\begin{equation}
 U = \left( x_{t^{*}} - x_{\text{target}} \right) \cdot \hat{u}   
\end{equation}

A negative value indicates that the trajectory stopped short of the target, while a positive value indicates an early overshoot followed by correction.

\subsubsection{Endpoint Error}

The endpoint error, denoted by \( E_{\text{final}} \), quantifies the final positional discrepancy at the end of a trial. It is computed as the Euclidean distance between the end-effector’s final position \( x_{\text{final}} \) and the target position \( x_{\text{target}} \):

\begin{equation}
 E_{\text{final}} = \left\| x_{\text{final}} - x_{\text{target}} \right\|   
\end{equation}

This metric reflects static accuracy and is independent of the trajectory’s transient dynamics such as overshoot or corrections.

\subsubsection{Spectral Arc Length}
Movement smoothness was quantified using the spectral arc length (SAL) metric \cite{6104119}. SAL is computed from the velocity profile’s power spectrum, where smoother trajectories exhibit smaller arc lengths. Lower SAL values indicate more erratic or segmented motion, whereas higher SAL values correspond to fluid movements.

\subsubsection{Economy of Motion}
In addition to smoothness, we also assessed the trajectory straightness, a metric that captures the efficiency of executed trajectories \cite{9594081, 11145187}. The path straightness index (PSI) is defined as the ratio between the straight line distance from the start point to the target and the actual path length traversed by the robot instrument tip. Formally, if $\ x_0$ and $\ x_T$ denote the start and end positions, and $\ x_t$ the trajectory samples, then

\begin{equation}
\text{PSI} = \frac{\lVert \ x_T - \ x_0 \rVert}{\sum_{t=1}^{T} \lVert \ x_t - \ x_{t-1} \rVert},
\end{equation}
where values closer to~1 indicate straighter, more economical trajectories, whereas smaller ratios reflect redundant or curved paths. Prior studies in telesurgery have shown that communication delay deteriorates economy of motion by increasing total path length and producing less direct movements \cite{mottetFittsLawTwodimensional1994, 258052}.

\subsubsection{Speed-Accuracy Tradeoff}

To evaluate whether our method improves user overall task performance under delayed condition, we define a compound performance metric that jointly captures spatial accuracy and temporal efficiency. Specifically, we compute a Combined Error-time (CET) \cite{oquendo_haptic_2024} score for each trial as:
\begin{equation}
    \text{CET} = \text{Endpoint Error} \times \text{Task Completion Time},
\end{equation}
\noindent where the \emph{Endpoint Error} is defined as the Euclidean distance between the final end-effector position and the goal location. Since our experiment includes reaching tasks with varying target distance, we normalize trial time using the Index of Difficulty (ID) from Fitts' Law:

\begin{equation}
    \text{ID} = \log_2\left(\frac{D}{W} + 1\right),
\end{equation}

\noindent where $D$ is the reaching distance and $W$ is the effective target width. The raw trial time $T_{\text{raw}}$ is then normalized as:

\begin{equation}
    T_{\text{norm}} = \frac{T_{\text{raw}}}{\text{ID}},
\end{equation}

\noindent yielding the normalized CET (nCET):

\begin{equation}
    \text{nCET} = \text{Endpoint Error} \times \left( \frac{T_{\text{raw}}}{\text{ID}} \right)
\end{equation}

This metric penalizes trials that are either inaccurate, slow (relative to task difficulty), or both. Lower values indicate better overall task efficiency. The use of Fitts' Law allows fair comparisons across reaching distances, ensuring that longer tasks are not inherently penalized for longer durations.
\subsubsection{Cognitive Load}
Subjective cognitive and physical workload were assessed using the NASA--TLX questionnaire. 
After each major experimental block, participants rated six dimensions: mental demand, physical demand, temporal demand, perceived performance, effort, and frustration. 
Each dimension was scored on a 0--20 scale, yielding a total workload score in the range 0--120. All dimensions were equally weighted when computing overall workload.

\vspace{-1em}
\subsection{Data Processing}
The experimental dataset was obtained from 20 participants performing a teleoperated reaching task under varying delay and assistance conditions. Each trial produced synchronous recordings of operator hand motion via the haptic device and the corresponding da Vinci PSM end-effector trajectory, sampled uniformly at 200\,Hz. On average, each subject contributed approximately six hours of motion data. Velocity profiles were derived by differentiating consecutive position samples from the optical tracker and low-pass filtering them using a fourth-order Butterworth filter with an 8Hz cutoff frequency \cite{RACZ2021102974}.

\vspace{-1em}
\subsection{Statistical Analysis}
We fitted linear mixed-effects models separately for each primary metric. Fixed effects included assistance paradigm, temporal delay, movement distance, and movement direction.
% were included to directly probe our hypotheses regarding the impact of latency and assistance. 
In SimOnly, assistance paradigm was modeled with two levels, with and without assistance. In Sim2Real, it was modeled with three levels, personalized assistance, generalized assistance, and no assistance. Direction was modeled as a two-level factor. Movements directed proximally toward the operator’s body midline were classified as inward (adduction), whereas movements directed away from the midline were classified as outward (abduction). Four-, three-, and two-way interactions between assistance paradigm, delay, direction and distance were modeled. A user's baseline ability was modeled using a participant's mean of that metric under no delay, for a given distance and direction. 

A random intercept for the user, and a geometry-matched 0\,ms baseline covariate (each participant’s mean in the same distance and direction), were included to model individual ability. In Wilkinson notation:
\begin{equation*}
\begin{aligned}
\text{metric} \sim {}&
\text{distance}\times\text{delay}\times\text{assistance}\times\text{direction}\\
& {}+ \mathrm{baseline} + (1\mid\text{user}),
\end{aligned}
\end{equation*}
where baseline denotes each participant’s geometry-matched baseline performance at 0\,ms. While the results of 4-way interactions are difficult to interpret, including these terms allowed modeling of specific shifts in motor behavior that might arise under specific task geometry conditions.

Type-III tests were used for omnibus inference on interaction terms, Satterthwaite degrees of freedom for p-values, and Tukey-adjusted estimated marginal means for simple-effects contrasts within each condition. 
%Because at 0 ms only Unassisted trials exist in our design, assistance contrasts are not estimated at 0 ms; the 0 ms information is used solely via the baseline covariate.

Residuals of fitted models were inspected for non-normality and non-constant variance using Q–Q plots and histograms. When required, Box–Cox transformations were used to improve distributional symmetry and stabilize variance. For the final analyses, Box-Cox transformations (parameterized by $\lambda$) were applied to all performance metrics except SAL in both the SimOnly and Sim2Real models. The estimated $\lambda$ values were as follows. For SimOnly, overshoot ($\lambda=0.22$), endpoint error ($\lambda=0.061$), PSI ($\lambda=1.11$), and nCET ($\lambda=-0.02$). For Sim2Real, overshoot ($\lambda=0.30$), endpoint error ($\lambda=0.18$), PSI ($\lambda=0.51$), and nCET ($\lambda=0.10$).

All statistical analyses were performed in R (v4.5.0), using the \texttt{lme4}, \texttt{lmerTest}, and \texttt{emmeans} packages. Estimated marginal means were obtained using \texttt{emmeans}. Pairwise contrasts between task states were evaluated with Tukey correction to control for multiple comparisons. 

\vspace{-1em}
\subsection{Empirical Delay-conditioned Gain Models}

To obtain a continuous model of the appropriate motion scaling compensation gain at a given delay in each reach direction, we fitted generalized additive models (GAMs) to the estimated gains. Because gain values were derived by normalizing overshoot against displacement, distance was excluded as a predictor to avoid redundancy and overfitting. Separate models were fitted for each movement direction, with delay as the sole predictor:
\[
G(\delta \mid \theta) = \beta_0 + f(\delta),
\]
where $\delta$ denotes temporal delay and $\theta$ indexes movement direction. The smooth term $f(\delta)$ was modeled using a spline basis of order four, which provided sufficient flexibility to capture nonlinear delay dependence while preventing overfitting. This procedure yielded direction-specific gain curves that interpolated across discrete delay conditions and reduced trial-level variability.

\section{RESULTS}

\begin{figure}[!t]
\centering
\includegraphics[width=0.45\textwidth]{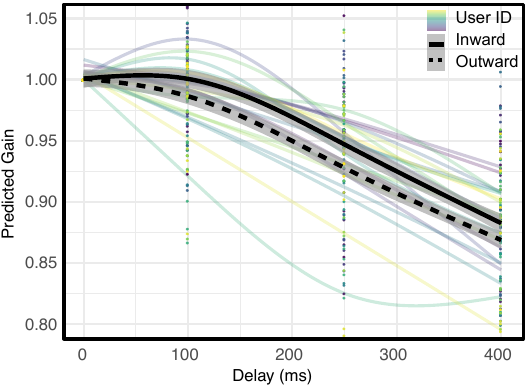}
\caption{Personalized and generic gains modeled as a function of temporal delay. Generic gains were obtained by averaging across participants.}
\label{fig:estimation of gain}
\end{figure}

Based on the data collected in the Gain ID block of the SimOnly task, we fitted the delay-conditioned gain models as shown in Fig.\,\ref{fig:estimation of gain}. Overall most users displayed a consistent gain-delay relationship where the required gain gradually decreases with longer delays. While some users did display hypometria at the 100\,ms delay, resulting personalized gains above unity, the generalized gain shows that on average, unity gain can be maintained up to 100\,ms.

The results of selected ANOVA analyses are summarized in Table~\ref{tab:anova_all}. For focus and interpretability we report and discuss all the main effects and two-way interaction effects associated with assistance. Full ANOVA details are provided in the supplementary materials. 

\vspace{-1em}
\subsection{Simulation Only Task}

\begin{figure*}[!t]
  \centering
  \includegraphics[width=\textwidth]{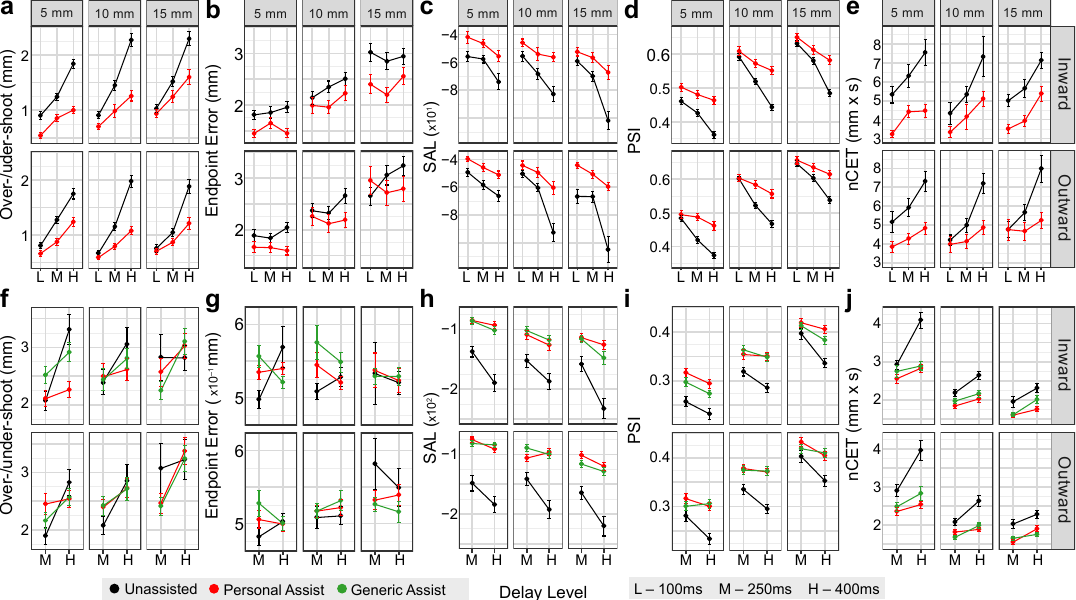} 
  \caption{All task performance metrics as a function of temporal delay, assist state, and target distance. 
Error bars indicate $\pm$1 SEM.}
  \label{fig:performance_metrics}
\end{figure*}

% \begin{figure*}[!t]
%   \centering
%   \includegraphics[width=\textwidth]{figures/s2r_2x3.pdf} 
%   \caption{All task performance metrics as a function of temporal delay, assist state, and target distance. 
% Error bars indicate $\pm$1 SEM.}
%   \label{fig:performance_metrics}
% \end{figure*}

% The effects of delay and assistance on spatial accuracy as defined by initial reach error (i.e. over and undershoot magnitude) and endpoint error are summarized qualitatively in Fig.\,\ref{fig:performance_metrics}a and c, respectively.  

% Across all metrics, the presence of assistance reduced error relative to the unassisted state. Taken together, these results suggest that while performance inevitably deteriorated with increasing delay, assistance systematically reduced the impact of latency on accuracy, smoothness, and straightness. Importantly, the advantage of assistance seemed to emerge most clearly at longer delays, and was consistent across movement directions, highlighting its robustness. Yet absolute performance still degraded with increasing delay, indicating that assistance mitigates but does not abolish delay-related impairments. In the following subsections, we provide a detailed report of our findings for each computed performance metric.

\subsubsection{Initial Reaching Error} 
Assistance, delay, and direction significantly predicted initial reaching errors. As seen in Fig. \ref{fig:performance_metrics}a, assistance resulted in significantly lower over-/under-shoot relative to unassisted trials broadly across delay, distance, and direction. Higher delays resulted in increased error across different directions and distances, regardless of assistance. Yet, there was a significant interaction between delay and assistance, indicating increasing accuracy benefit with assistance under more severe latency. Post-hoc pairwise comparisons revealed significant and increasing error reductions with personalized assistance relative to the unassisted condition at all delays (100\,ms: $\Delta\approx-0.035$, $p<0.001$; 250\,ms:  $\Delta\approx-0.068$, $p<0.001$; 400\,ms:  $\Delta\approx-0.122$, $p<0.001$). 
%This suggests a growing benefit of assistance under more severe latency. (LEAVE THIS FOR DISCUSSION)

% Initial reaching errors increased with longer delays and varied with movement direction, as shown in Table~\ref{table:so_anova}; the effect of target distance showed no significant difference. Assisted trials yielded significantly lower overshoot relative to the unassisted conditions ($p<0.001$), and undershoot errors showed a similar reduction. Both errors tended to increase markedly with longer communication delays ($p<0.001$), varied significantly across movement directions ($p<0.001$), and tended to increase with larger reaching distance (Fig.~\ref{fig:performance_metrics}a). Interactions indicated that the assistance effect varied with distance, delay, and direction. Assistance$\times$Distance was significant ($p<0.01$), Assistance$\times$Delay was the largest interaction ($p<0.001$), and Assistance$\times$Direction was weaker but reliable ($p<0.05$).

\begin{table*}[htbp]
\scriptsize
\caption{Selected Type-III ANOVA results for all SimOnly and Sim2Real performance metrics}
\label{tab:anova_all}
\centering
\begin{adjustbox}{width=\textwidth}
\begin{tabular}{lccccccccccccccc}
\toprule
\textbf{Effect} &
\multicolumn{3}{c}{\textbf{Over-/under-shoot}} &
\multicolumn{3}{c}{\textbf{Endpoint Error}} &
\multicolumn{3}{c}{\textbf{SAL}} &
\multicolumn{3}{c}{\textbf{PSI}} &
\multicolumn{3}{c}{\textbf{nCET}} \\
\cmidrule(lr){2-4}\cmidrule(lr){5-7}\cmidrule(lr){8-10}\cmidrule(lr){11-13}\cmidrule(lr){14-16}
& F & Pr($>$F) & Sig
& F & Pr($>$F) & Sig
& F & Pr($>$F) & Sig
& F & Pr($>$F) & Sig
& F & Pr($>$F) & Sig \\
\midrule
\multicolumn{8}{l}{\textit{SimOnly}}\\
Asst      & 273.43 & $<0.001$ & *** &  75.60 & $<0.001$ & *** & 226.38 & $<0.001$ & *** & 334.89 & $<0.001$ & *** & 218.94 & $<0.001$ & *** \\
Delay     & 467.05 & $<0.001$ & *** &  14.66 & $<0.001$ & *** & 211.23 & $<0.001$ & *** & 285.29 & $<0.001$ & *** & 208.24 & $<0.001$ & *** \\
Distance  &   1.02 & 0.35     &     &  45.51 & $<0.001$ & *** &  80.85 & $<0.001$ & *** & 124.47 & $<0.001$ & *** &  20.91 & $<0.001$ & *** \\
Direction &  14.73 & $<0.001$ & *** &   4.04 & 0.04     & *   &   9.90 & 0.002    & **  &  43.76 & $<0.001$ & *** &   1.33 & 0.25     &     \\
Baseline  & 252.45 & $<0.001$ & *** & 1290.11 & $<0.001$ & *** &  34.17 & $<0.001$ & *** & 829.62 & $<0.001$ & *** & 1287.68 & $<0.001$ & *** \\
Asst:Delay    & 31.16 & $<0.001$ & *** &  1.87 & 0.13    &     &   3.54 & 0.03     & *   &  57.16 & $<0.001$ & *** &   4.38 & 0.01     & *   \\
Asst:Distance &  6.70 & $<0.01$  & **  &  0.34 & 0.71    &     &   0.25 & 0.78     &     &   2.75 & 0.06     & .   &   0.98 & 0.37     &     \\
Asst:Direction&  5.78 & 0.01     & *   &  6.83 & 0.01    & *   &   0.04 & 0.84     &     &   2.83 & 0.09     & .   &   6.84 & $<0.01$  & **  \\
\midrule
\multicolumn{8}{l}{\textit{Sim2Real}}\\
Asst      &  4.22 & 0.01     & *   &   3.25 & 0.04     & *   & 330.12 & $<0.001$ & *** & 146.22 & $<0.001$ & *** & 129.38 & $<0.001$ & *** \\
Delay     & 73.81 & $<0.001$ & *** &   0.05 & 0.83     &     & 106.97 & $<0.001$ & *** & 83.53 & $<0.001$ & *** & 150.76 & $<0.001$ & *** \\
Distance  &  2.74 & 0.06     & .   &   3.68 & 0.03     & *   &  62.71 & $<0.001$ & *** & 115.50 & $<0.001$ & *** &  79.86 & $<0.001$ & *** \\
Direction &  0.03 & 0.87     &     &   3.67 & 0.06     & .   &  10.74 & $<0.01$  & **  &  21.60 & $<0.001$ & *** &  14.69 & $<0.001$ & *** \\
Baseline  &  6.75 & 0.01     & *   & 240.94 & $<0.001$ & *** &  12.08 & $<0.001$ & *** & 115.58 & $<0.001$ & *** &  81.01 & $<0.001$ & *** \\
Asst:Delay    & 0.49 & 0.61  &     &   4.89 & 0.01     & **  &   8.92 & $<0.001$ & *** &  13.87 & $<0.001$ & *** &  10.62 & $<0.001$ & *** \\
Asst:Distance & 0.27 & 0.90  &     &   0.71 & 0.58     &     &   5.73 & $<0.001$ & *** &   3.03 & 0.02     & *   &   1.21 & 0.31     &     \\
Asst:Direction& 4.65 & 0.01  & *   &   1.64 & 0.19     &     &   0.45 & 0.64     &     &   0.66 & 0.52     &     &   2.45 & 0.09     & .   \\
\bottomrule
\end{tabular}
\end{adjustbox}
\end{table*}

\subsubsection{Endpoint Error} 
All four main effects significantly predicted endpoint error. As with initial reaching error, assistance was broadly beneficial at reducing endpoint error (Fig.~\ref{fig:performance_metrics}b). However, unlike in initial reaching error, the benefit of assistance did not increase relative to no assistance as delay increased. Instead, there was a significant interaction of assistance with reach direction, in which assistance was more beneficial, and its effect more consistent, in the inward reaching direction, relative to no assistance.

%Analysis of endpoint error revealed parallel main effects. Controlling for each participant’s baseline endpoint error ($p<0.001$), assistance reduced error overall ($p<0.001$). Endpoint error also varied with distance ($p<0.001$), movement direction ($p<0.05$), and delay ($p<0.001$). Although interactions with distance and delay were not significant (Table.~\ref{table:so_anova}), a significant Assist,$\times$, Direction effect emerged ($p<0.05$), indicating that the benefit of assistance depended on direction of movement. As shown in Fig.~\ref{fig:performance_metrics}c, inward movements benefited more from assistance than outward movements.

\subsubsection{Movement Smoothness} Assistance, delay and distance significantly predicted movement smoothness. Assistance broadly increased movement smoothness across tested conditions (Fig.~\ref{fig:performance_metrics}c). Similar to initial endpoint error, there was a significant interaction between delay and assistance, indicating increasing smoothness benefit with assistance under more severe latency. Compared to the unassisted condition, personalized assistance reduced SAL values at all delay conditions (100\,ms: $\Delta\approx6$, $p<0.001$; 250\,ms:  $\Delta\approx7.8$, $p<0.001$; 400\,ms:  $\Delta\approx13.9$, $p<0.001$). 

% Smoothness decreased with increasing delay (Fig.~\ref{fig:sal}). SAL values revealed significant main effects of Assist ($p<0.001$), Distance ($p<0.001$), Delay ($p<0.001$), and Direction ($p<0.001$) in the mixed-effects model. Assistance $\times$ Delay was significant ($p<0.001$), indicating delay-dependent modulation of the assistance benefit, while other Assist-involving interactions were not significant. Compared to the unassisted condition, personalized assistance reduced SAL values by approximately 6.0 units at 100~ms, 7.8 units at 250~ms, and 13.9 units at 400~ms delay (all $p<0.001$). These reductions indicate that under assistance movement smoothness improved more under longer delays
% , suggesting that assistance was especially effective under greater temporal challenge 
%(See Appendix~\ref{tab:pairwise_simonly}).

\subsubsection{Movement Economy} All four main effects significantly predicted movement economy. Assistance improved movement economy generally across all experimental factors, enabling straighter reaches (Fig.~\ref{fig:performance_metrics}d). Assistance had significant two-way interactions with delay. Notably, assistance improved economy of motion, relative to no assistance, under more severe latency. Post-hoc comparisons across non-zero delays showed a graded assistance benefit, small at 100~ms ($\Delta\approx0.016$, $p<0.001$), larger at 250~ms ($\Delta\approx0.047$, $p<0.001$), and largest at 400~ms ($\Delta\approx0.086$, $p<0.001$).

% Movement economy was impaired with longer delays and varies with distance and direction in the SimOnly environment. Controlling for baseline straightness ($p<0.001$), assistance improved straightness overall ($p<0.001$). Distance, Delay, and Direction showed strong main effects (Distance: $p<0.001$; Delay: $p<0.001$; Direction: $p<0.001$). Moreover, Assistance$\times$Delay and Assistance$\times$Distance were significant ($p<0.001$; $p<0.05$), whereas other high-order interactions were not significant. Post-hoc comparisons across non-zero delays showed a graded assistance benefit, small at 100~ms ($\Delta\approx-0.016$, $p<0.001$), larger at 250~ms ($\Delta\approx0.047$, $p<0.001$), and largest at 400~ms ($\Delta\approx0.086$, $p<0.001$). Overall, assistance enabled better movement straightness with increasing delay across all directions, with modest distance-dependent modulation via the Assist$\times$Distance interaction.

\subsubsection{Combined Error-time} 
\begin{figure}[!t]
\centering
\includegraphics[width=0.45\textwidth]{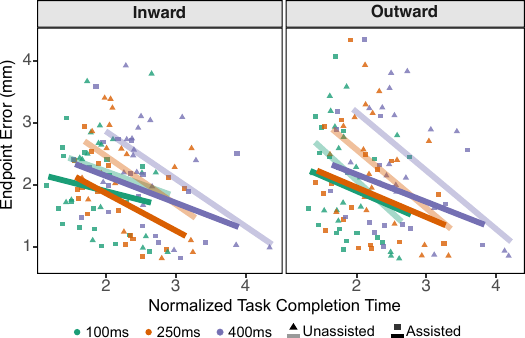}
\caption{Speed-accuracy curves from end-point error versus normalized completion time in the SimOnly scenario. 
% This downward displacement reflects the overall benefit of the assistive Assist in enhancing user performance under delayed teleoperation.
}
\label{fig:errorxtime}
\end{figure}

Assistance, delay, and distance significantly predicted normalized CET.
% (all $p<0.001$), indicating that each factor individually influenced task efficiency. 
% Although Direction showed no significant main effect,
There was a significant three-way interaction between assistance, delay, and direction, and two-way interaction of assistance with delay and direction. Post-hoc pairwise comparisons for the nCET metric showed significant improvements with personalized assistance relative to the unassisted condition across all delays (Fig.~\ref{fig:performance_metrics}e). The magnitude of improvement under assistance grew systematically with delay. This is also illustrated by the Error$\times$Time plot in Fig.\,\ref{fig:errorxtime}, where assistance clearly shifts the curves downward and to the left, implying a better speed-accuracy trade off (100\,ms: $\Delta\approx0.129$, $p<0.001$; 250\,ms: $\Delta\approx0.184$, $p<0.001$; 400\,ms: $\Delta\approx0.224$, $p<0.001$).
% , indicating that assistance
% became increasingly beneficial under more severe latency as shown in
% Fig.~\ref{fig:errorxtime}.

% Across all evaluated metrics in the SimOnly environment, personalized assistance significantly improved performance under communication delay. For most metrics, the benefit of assistance scaled with increasing latency, confirming the intended design of our adaptive method. However, only a subset of metrics exhibited significant interactions with spatial factors such as target distance or movement direction. Notably, Assist$\times$Distance effects emerged for initial reaching error and path straightness, while Assist$\times$Direction effects were observed for endpoint error and task efficiency.

% These findings suggest that while delay imposes a consistent challenge across movement tasks, making it a dominant modulator of performance, spatial interactions with assistance emerge only when the metric itself is sensitive to trajectory shape, movement timing, or directional asymmetries. Thus, the efficacy of assistance under delay is broadly generalizable, but its spatial specificity is metric-dependent, reflecting intrinsic differences in how each measure captures motor performance.

\begin{figure}[!t]
\includegraphics[width=0.45\textwidth]{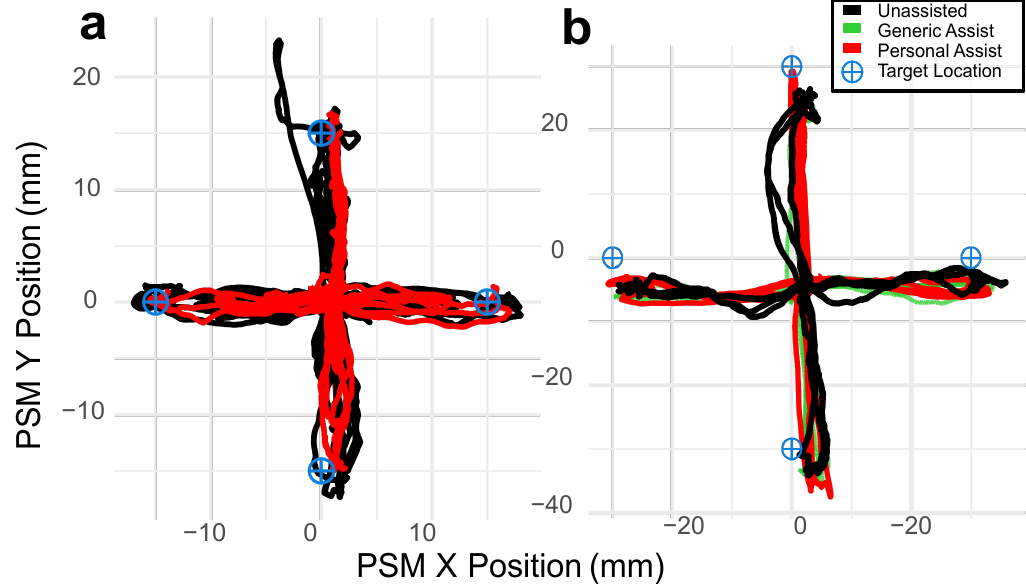}
  \caption{Reach paths under assisted vs.\ unassisted conditions for (a) simulation-only and (b) sim-to-real conditions.}
  \label{fig:eco_motion}
\end{figure}

\subsubsection{Perceived Workload} There was a significant main effect of assistance on TLX scores
($F(1,197)=80.11$, $p<0.0001$). Estimated marginal means indicated that Personalized assistance yielded lower TLX scores ($M=46.1$, $SE=4.8$) compared with the Unassisted condition ($M=50.2$, $SE=4.8$). 
Planned pairwise comparisons confirmed that this difference was significant 
(mean difference $=-4.17$, $SE=0.47$, $t(197)=-8.95$, $p<0.0001$). 
These findings demonstrate that, in the SimOnly setting, individualized scaling substantially reduced subjective workload relative to unassisted control. 

\subsection{Sim-to-Real transfer Task}

\subsubsection{Initial Reaching Error}

Assistance and delay significantly predicted initial reach error, consistent with findings in SimOnly. As seen qualitatively in Fig.\,\ref{fig:performance_metrics}f on average, assistance broadly reduced error across all experimental parameters. There were significant two-way interaction effects of assistance with delay, distance, and direction. However, unlike in SimOnly, there was no consistent, increased benefit of assistance at higher delays. Instead, assistance reduced error at higher delays only for 5\,mm and 10\,mm reach distances; beyond that, assistance did not demonstrate beneficial effects. Similar to SimOnly, inward reaching saw greater benefit from assistance compared to outward reaching. Pre-planned orthogonal contrasts showed that significant error reductions for assisted vs. unassisted ($p<0.05$), and for personalized vs. generalized assistance ($p<0.01$), were confined to the inward reaching only.

% Initial reaching errors in the physical system were strongly modulated by communication delay, consistent with simulation findings. While assistance improved performance overall, its benefit was direction-dependent, most pronounced during inward (adduction) movements. Significant Assist × Direction and Distance × Direction interactions suggest that assistance effects are shaped not only by latency but also by task geometry and movement visibility. 

% Interestingly, personalized scaling outperformed generic scaling in reducing overshoot during inward movements, indicating that tailored strategies may offer additional benefits in Sim2Real transitions where spatial dynamics are directionally asymmetric. In contrast, no significant improvement was observed for outward movements, highlighting that the efficacy of assistance is not uniformly distributed across spatial directions.

\subsubsection{Endpoint Error}
% Type-III mixed-effects ANOVA on endpoint error showed main effects of Assist ($p\approx0.04$), Distance ($p\approx0.03$). An Assist$\times$Delay was significant ($p\approx0.01$), and a two-way Distance$\times$Direction interaction reached significance ($p<0.001$); all other interactions were non-significant. Taken together, assistance-related differences in terminal accuracy were not global but context-dependent, emerging only in specific Delay$\times$Distance configurations (see Appendix for post-hoc details).

Assistance and distance significantly predicted endpoint error. However, assistance did not result in consistent error reduction compared to no assistance across experimental conditions (Fig.~\ref{fig:performance_metrics}g). There was a two-way interaction between assistance and delay, which can be seen qualitatively in the crossed trends of the different assistance conditions from 250\,ms to 400\,ms delay. Pre-planned contrasts confirmed that the only significant effect was actually a decrease in accuracy due to assistance at 250\,ms delay ($p < 0.001$). 

% Terminal accuracy was modestly improved by assistance and was sensitive to both delay and distance. However, unlike overshoot, the effect of assistance on endpoint error was more selective, emerging primarily in specific delay-distance configurations. The lack of consistent interaction with direction suggests that endpoint error may be less susceptible to visuomotor asymmetries but more affected by temporal uncertainty and reach extent. This divergence highlights a key contrast with overshoot, reinforcing that not all metrics respond uniformly to assistance.

% \begin{figure}
% \centering
% \includegraphics[width=0.45\textwidth]{figures/s2r_sal_inout.pdf}
% \caption{Trajectory smoothness summarized across delay, direction, distance, and task state. Less negative SAL indicates smoother trajectories. Assisted trials consistently shift the distribution toward smoother motion relative to unassisted.}
% \label{fig:sal}
% \end{figure}

\subsubsection{Movement Smoothness} 
% Smoothness (Fig.~\ref{fig:sal}) showed significant main effects of Assist, Delay, Distance (all $p<0.001$), and Direction ($p<0.01$). Importantly, a significant Assist x Delay and Assist$times$Distance interaction was observed ($p<0.001$), indicating that the effect of assistance depended on temporal latency. In contrast, Assist × Direction and higher-order interactions were not significant ($p>0.6$), suggesting that the benefits of assistance varied depending on delay and reaching distance without directional specificity. Post-hoc contrasts confirmed that assisted trials consistently outperformed the unassisted condition across all tested delays and distances (all $p<0.001$). The magnitude of this benefit was evident in both lateral and longitudinal directions, indicating a robust and generalizable effect of assistance. By contrast, no significant differences were found between generic and personal assistance ($p>0.1$ across all conditions), suggesting that both strategies yielded comparable improvements in smoothness. 

All four experiment factors significantly predicted movement smoothness. While on average smoothness declined with higher delay (Fig.~\ref{fig:performance_metrics}h), assistance resulted in improved smoothness compared to no assistance. Assistance had significant two-way interactions with delay and direction. A pre-planned contrast at different delay levels revealed a significant increase in smoothness at all delay levels due to assistance (250\,ms: $p<0.001$, 400\,ms: $p<0.001$).

\subsubsection{Movement Economy} All four fixed effects significantly predicted movement economy. Assistance had a significant interaction with delay, in which assistance under longer delays resulted in more improvement in movement economy relative to no assistance (Fig.\,\ref{fig:performance_metrics}i). Pre-planned contrast testing confirmed significance between assistance vs. no assistance over the two delays levels (250\,ms: $p<0.001$, 400\,ms: $p<0.001$). However, the benefit of generalized versus personalized assistance was only significant at 400\,ms delay.

\subsubsection{Combined Error-Time}  
All four fixed effects significantly predicted normalized CET. There was a significant interaction between assistance and delay, with pre-planned contrasts showing that assistance produced greater speed-accuracy benefits relative to no assistance at all delay levels (250\,ms: $p<0.001$, 400\,ms: $p<0.001$). Comparing the two forms of assistance, personalized assistance resulted in significantly lower nCET, compared to generalized assistance only at the higher delay level (400\,ms: $p < 0.05$).

% The integrated Error$/times$Time metric exhibited a similar trend to PSI, captured overall efficiency and revealed broad benefits from assistance. These gains increased with delay and interacted with direction, suggesting that performance under spatially demanding conditions (e.g., inward or long-reaching movements) benefited more from assistance under latency (See Appendix~\ref{tab:contrast_s2r}). 

\subsubsection{Perceived Workload} A significant main effect of assistance on TLX scores was observed 
($F(2,304)=6.09$, $p=0.0026$). However, contrasts revealed no significant difference between Unassisted and Assisted trials 
(mean difference $=-0.47$, $SE=0.88$, $t(304)=-0.54$, $p=0.592$), implying no consistent effect across different experimental conditions. 

%One possible explanation is that, under assisted conditions, the effective gain was continuously adjusted across different task settings. Such changes may have disrupted the sensorimotor calibration that participants had already established during unassisted trials, leading to a higher subjective workload despite objective improvements in performance.

However, Personalized assistance produced significantly lower TLX scores than the Generic gain 
(mean difference $=-3.50$, $SE=1.01$, $t(304)=-3.45$, $p=0.0013$). 
These results suggest that, in the Sim2Real setting, overall assistance did not alter subjective workload relative to unassisted control. However, personalization conferred a distinct benefit by reducing workload compared with a generic gain. 

% Across all metrics, assistance improved user performance in the presence of communication delay, but the extent and nature of this improvement varied depending on the metric and the specific task context. Delay emerged as the most consistent modulator of performance; assistance was particularly effective under high-latency conditions. However, not all performance dimensions responded to assistance in the same way. Initial reaching and endpoint error, which reflect spatial accuracy, showed context-dependent improvements, especially in inward (adduction) movements, suggesting that direction-specific biomechanical or perceptual demands can modulate the effectiveness of assistance. In contrast, smoothness and straightness were more uniformly enhanced by assistance across directions, with effects primarily shaped by delay and reach distance. These metrics appear to capture more global aspects of motor efficiency and were less sensitive to directional asymmetries. The CET metric further highlighted the combined influence of spatial and temporal demands, revealing complex interactions among delay, direction, and distance. 

% Importantly, the alignment between simulation and physical results supports the successful sim-to-real transfer of our proposed scaling framework, particularly under challenging delay conditions. This suggests that the effectiveness of assistance strategies depends not only on algorithmic design but also on the behavioral dimension being targeted, and that generalizable benefits can be achieved when these factors are properly matched.

\section{Discussion}

% This study examined how temporal delay alters movement execution in telemanipulation and evaluated the extent to which motion scaling can mitigate delay-induced impairments. Across both simulation and real-world settings, delay consistently increased overshoot, endpoint error, and reduced trajectory smoothness and efficiency. 

The benefits of sub-unity motion scaling were strongly and consistently seen across all metrics. The benefits were, on the whole, more pronounced at higher delay and distance levels, supporting prior work showing that this form of feedback is a practical delay mitigation strategy. The benefit for endpoint error was less pronounced and consistent compared to initial reaching error, especially for higher delays and longer reaches (15\,mm). This could have been due to the fact that subsequent sub-movements after the initial reach could have relied more on feedback control, especially in those particular experimental conditions.

The benefit of assistance on accuracy was modulated by the reaching direction, with inward, i.e. proximal, reaching directions showing consistent increases in accuracy. These were particularly noticeable for the endpoint error in SimOnly and for the initial reaching error in Sim2Real (Fig.,\ref{fig:performance_metrics}b and g, respectively). From a biomechanical perspective, arm proprioception is more accurate for hand locations closer to the body, leading to the directional modulation of assistance benefit \cite{christinat.fuentesWhereYourArm2010,rincon-gonzalezProprioceptiveMapArm2011,chuaTaskDynamicsPrior2020}. Given this increased reliability, subjects may have depended less on delayed visual feedback when reaching inward, relying more on the feedforward model that the assistance parameters were tuned to. Conversely, when proprioceptive uncertainty is higher (e.g., for more distal targets), subjects may have weighted the delayed visual feedback more heavily, reducing the effect of assistance. Practically, telesurgery frequently involves clutching to keep the hands centered in the workspace \cite{clutch2neutral, kazanzidesOpenSourceResearchKit2014, reileyReviewMethodsObjective2011}. Thus, degradation of beneficial assistance is likely to be limited.

Unlike initial studies of motion scaling, in which hand engineered, i.e. non-data-driven gains were used, resulting in a notable speed accuracy tradeoff \cite{richterMotionScalingSolutions2019}, our empirical approach yielded better speed-accuracy, and supports later findings that optimal scaling results in higher task throughput \cite{luAdaptiveControlTime2022}.

%\cite{christinat.fuentesWhereYourArm2010, kwakkelStandardizedMeasurementQuality2019, rincon-gonzalezProprioceptiveMapArm2011, tsayPositionSenseHuman2016, beerDeficitsCoordinationMultijoint2000}.

%Prior studies have shown that arm movements performed near the limits of joint range are less precise and more variable due to reduced mechanical advantage and sensory resolution 

\vspace{-1em}
\subsection{Simulation to Real Transfer}
Assistance parameters learned in simulation demonstrated specific transferability to the real task. Initial reach error, movement smoothness and economy, and speed-accuracy were improved under assistance, though the benefit was reduced and less consistent, and was eliminated for initial reaching error. Assistance did not improve end-point error in the real task. 

The simulation to real gap likely arose from several experimental design choices. First, compared to shape matching of the ring, the peg transfer task has less restrictive goal: the task is considered successful if the ring is transferred onto the peg, regardless of whether the peg is exactly centered on the ring's inner hole. This meant that users were more likely to merely satisfy the task completion requirement, as opposed to maximizing endpoint accuracy, thus introducing more variability. Second, to fully isolate the scaling factor in simulation, we opted to present a planar view of the task to the user to remove visual uncertainty. In the real task, we reverted to the more realistic perspective viewpoint, reintroducing the depth uncertainty. This uncertainty could have made user move more cautiously, potentially relying more on closed-loop control via delayed visual feedback. Such a shift in control strategy could thus limit the benefits of assistance at higher delay levels and longer reaches.

\vspace{-1em}
\subsection{Personalized vs. Generic Assistance}
Personalized gains derived from simulation data improved performance for some participants but did not produce consistent advantages over a generic gain when averaged across the entire participant group. Promisingly, there were specific aspects of movement that benefited from personalization in conditions of reduced perceptual uncertainty.

For initial reaching error, personalization was most effective during inward movements and at low visual delays, where proprioceptive estimates were likely more reliable and reliance on delayed visual feedback was lower. At higher delays, personalization improved participants' speed–accuracy metric. Because speed–accuracy reflects a spatio-temporal tradeoff rather than a purely positional measure, it likely captured performance improvements that emerge from assistance when participants adjusted the temporal dimensions of their movement strategies under longer delays.

Outside of these specific use cases, the generic gain can provide a strong principled starting point for new users. Fine tuning of the gain can be then used to obtain highly optimized performance for a particular applications, e.g., tasks in which more proximal reaches needed. 

% This view aligns with the observation that users adjust their motor strategies under delay, and that scaling parameters tend to converge once this adaptation stabilizes. The practical significance of the generic gain is therefore twofold: it not only enables clustering and efficient initialization at a population baseline, but also highlights cases where experienced operators, such as surgeon, diverge substantially from the mean. In such instances, a personalized or adaptive extension of the framework may yield further gains in task efficiency beyond the generic baseline.

\vspace{-1em}
\subsection{Limitations and Future Work}
The tasks studied here were restricted to translational reaching and peg transfer, with wrist rotations locked to reduce confounds. Naive orientation scaling and clutching of orientations is unintuitive, and further development is needed to realize sub-unity assistive scaling approaches for those degrees of freedom. Furthermore, more complex manipulations may also amplify individual differences and could alter the relative advantage of personalized strategies \cite{kroemerReviewRobotLearning2020, tsujiSurveyImitationLearning2025, SUOMALAINEN2022104224}. 

While effective, the direction and distance-conditioned gain fitting approach used required a time consuming calibration process. Future work should investigate integrating online parameter adaption approaches \cite{luAdaptiveControlTime2022} with direction- and distance-specific intent conditioning, in which contextual cues such as task phase, or movement orientation with intent estimation.

% The sample size, although larger than many prior teleoperation studies, still limits statistical power to detect subtle effects. The present findings should therefore be interpreted as a conservative estimate of the potential value of personalization, which may be more pronounced in dexterous manipulation tasks. Future work should incorporate equivalence testing and hierarchical modeling to better characterize the balance between population-level regularities and user-specific variability.

% \vspace{-1em}
% \subsection{Future Directions}
% The present findings point toward adaptive and context-aware compensation as a promising path forward. Online adaptation could refine scaling parameters as users interact with the system, while contextual signals such as gaze, task phase, or movement orientation could guide when and where compensation should be applied. Cluster-based approaches may also prove valuable by identifying subgroups of operators with similar delay sensitivities, thereby balancing generality with personalization. Taken together, these strategies can move beyond static calibration toward compensation that evolves with both the operator and the task environment.

\section{Conclusion}
This work investigated a principled framework for mitigating delay impairment in telemanipulation through the design and evaluation of personalized motion scaling. Controlled experiments demonstrated that motion scaling assistance improved operator performance, with increased benefit at longer delays. We show that under certain conditions, the fitted scaling gains exhibit both directional, and distance dependence. Our simulation-based parameter fits also showed generalization to real hardware, though with reduced benefits, particularly at a long delay and reach distance. We show that leveraging the population level averages provided an effective generic gain for new users, while also supporting further personalization in conditions where human feedforward control is needed but uncertain. 
% In this way, the method balances robustness and adaptability, offering a practical pathway toward delay compensation that is both scalable and user sensitive.

% At a broader level, the findings highlight that the effects of temporal delay are neither uniform across individuals nor across task geometries. Personalized scaling produced selective improvements where biomechanical constraints amplified delay-induced errors, whereas generic scaling was sufficient in most scenarios. This dual pattern underscores that while a generic baseline can provide an efficient starting point, meaningful gains emerge when compensation is tuned to the individual and the specific demands of the task.

% Our framework is complementary to existing interaction paradigms in telesurgery, such as clutching. 

% Delay compensation should be understood not as a static calibration but as a dynamic and context-aware process. Given our finding of direction and distance dependence, future combination of online parameter optimization should seek to incorporate these conditional variables to further enable fast, smooth and accurate telemanipulation.

% Embedding adaptive compensation directly into teleoperation systems could therefore enhance performance without requiring extensive retraining, supporting both novice and expert operators.

In conclusion, delay compensation should be understood not as a static calibration but as a dynamic and context-aware process. By coupling principled population initialization with adaptive refinement informed by user state and task context, teleoperation systems can achieve both reliability and flexibility. Future research should thus advance context-aware strategies that dynamically refine compensation parameters online. Such integration would move the field closer to resilient human–robot interaction, enabling safer and more efficient execution in surgical and other time-critical applications. 
% Such methods have the potential to improve the reliability and precision of remote surgical and telemanipulation systems in the presence of communication delays.

\section*{Acknowledgments}
% This work was supported in part by the Office of Naval Research (ONR).

We would like to thank Shuyuan Yang for his support in calibrating the MicronTracker system, which was essential for the experimental setup. We also thank Nicholas Zingale, Douglas Wajda, and other colleagues for helpful discussions and feedback on the human factors aspects of this work.

\bibliographystyle{IEEEtran}
\bibliography{reference_betterbib}

\end{document}